\documentclass[letterpaper]{article} 
\usepackage{aaai2027}  
\usepackage[hyphens]{url}  
\usepackage{graphicx} 
\usepackage{natbib}  
\usepackage{caption} 
\usepackage{algorithm}
\usepackage{algorithmic}
\usepackage{amsmath}
\usepackage{amssymb}
\usepackage{booktabs}
\usepackage[table]{xcolor}
\usepackage{multirow} 
\usepackage{xcolor}
\definecolor{diffgreen}{RGB}{0, 130, 0}
\definecolor{diffred}{RGB}{130, 0, 0}

\usepackage{newfloat}
\usepackage{listings}
\DeclareCaptionStyle{ruled}{labelfont=normalfont,labelsep=colon,strut=off} 
\floatstyle{ruled}
\newfloat{listing}{tb}{lst}{}
\floatname{listing}{Listing}

\usepackage{booktabs}

\title{From Space to Math: Teaching MLLMs to Abstract
\\Spatial Reasoning as Mathematics}

\author{
    Li Ma,
    Yan Zhuang,
    Jiawen Deng\corresponding,
    Fuji Ren\corresponding
}

\affiliations{
    University of Electronic Science and Technology of China\\
    No. 2006, Xiyuan Ave, West Hi-Tech Zone, Chengdu, Sichuan, 611731, P.R. China\\
    \{202421080935, 202211081370\}@std.uestc.edu.cn,
    \{dengjw, renfuji\}@uestc.edu.cn
}

\begin{document}

\maketitle


\begin{abstract}
Spatial reasoning remains a persistent bottleneck of multimodal large language models (MLLMs). Existing methods, whether training-based or relying on external information sources, almost universally conduct reasoning in natural language, which loses numerical precision and entangles perception with reasoning. We observe that spatial reasoning problems are inherently mathematical, and that the resulting mathematical formulations are typically elementary for modern MLLMs. Motivated by this, we propose \textbf{MASIR}, a new reasoning paradigm that abstracts spatial problems into their mathematical form and solves them accordingly. To train this capability, we develop a staged SFT pipeline with an automated data construction process. GRPO follows to improve generalization, with general reasoning problems incorporated to preserve the model's broader capabilities. Extensive experiments demonstrate consistent and substantial improvements across diverse benchmarks, attaining state-of-the-art performance among open-source models across all benchmarks, and surpassing closed-source models on half of the benchmarks.
\end{abstract}


\begin{figure}[t]
\centering
\includegraphics[width=\columnwidth]{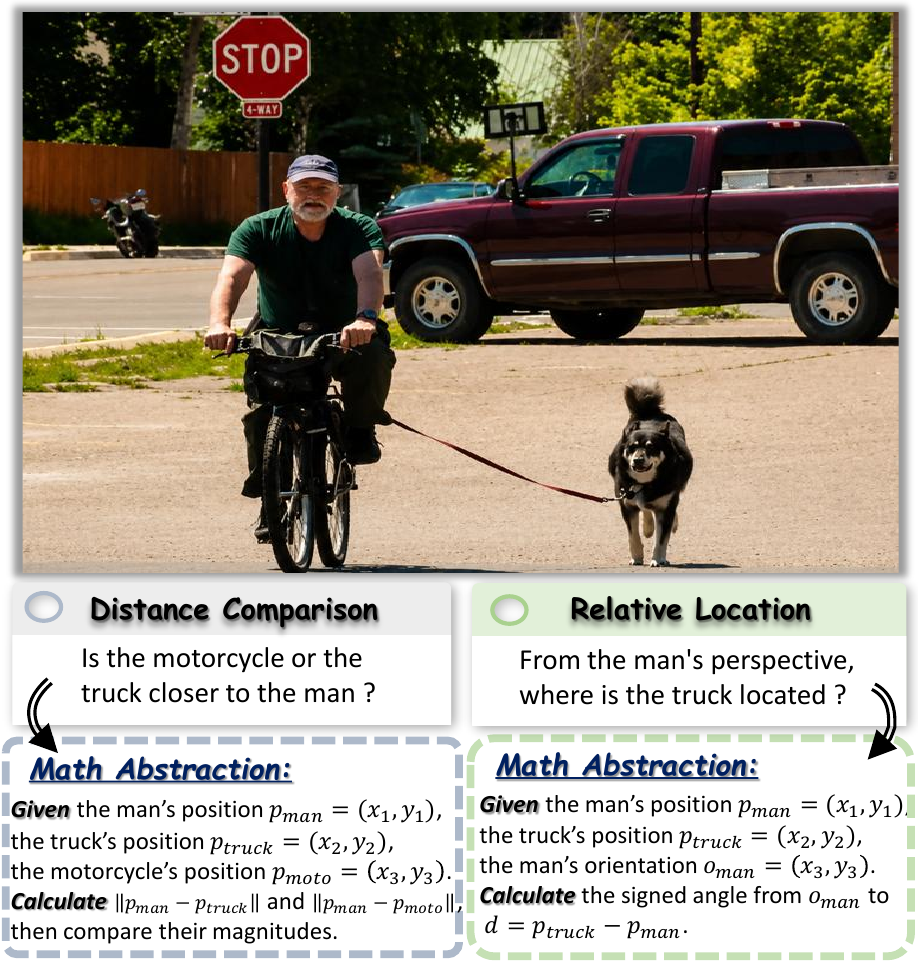}
\caption{Mathematical abstraction of two spatial reasoning tasks. Each query is converted into variables and geometric operations needed to obtain the answer.}
\label{fig:intro}
\end{figure}

\section{Introduction}

Recent multimodal large language models (MLLMs) exhibit a clear imbalance in reasoning capability. While performing exceptionally well on olympiad-level mathematics\cite{MathVista} and complex code generation tasks\cite{CodeGeneration}, they consistently underperform on basic spatial reasoning tasks such as distance estimation and directional relationship inference over real-world scenes. This limitation has direct consequences for embodied navigation, robotic manipulation, and autonomous driving, where accurate spatial understanding is a core requirement.

Existing approaches to this problem can be broadly categorized as training-based methods\cite{SpatialVLM,SpatialLadder} and methods that introduce external information sources\cite{SandboxVLM,SpatialPIN}. However, they share a common characteristic: reasoning is conducted entirely in natural language. This introduces two fundamental problems. First, natural language is coarse-grained and loses numerical precision. Describing a distance as ``\textit{slightly farther}'' discards the very information needed for a correct answer. Second, perception and reasoning become entangled within the same language stream, leading to confused reasoning and reduced efficiency. This also makes it difficult to determine where the failure originates.

This raises a natural question: \textit{can a new reasoning paradigm emerge to mitigate these limitations?} Two observations motivate our approach. First, spatial reasoning problems are inherently mathematical. As shown in Figure~\ref{fig:intro}, physical attributes such as orientations and locations can be expressed as variables, and the original query can be abstracted into a well-defined mathematical problem. Second, modern MLLMs have demonstrated strong mathematical reasoning ability. The mathematical problems abstracted from the original queries are typically elementary, which can be readily solved by current models.
Motivated by these observations, we propose \textbf{MASIR} (\textit{\textbf{M}athematical \textbf{A}bstraction for \textbf{S}pat\textbf{I}al \textbf{R}easoning}), a new reasoning paradigm that abstracts spatial reasoning problems into their inherent mathematical form and solves them accordingly.

Specifically, MASIR decomposes inference into three explicit steps: \textit{Problem Abstraction}, \textit{Spatial Perception}, and \textit{Mathematical Computation}. 
The first step abstracts the original spatial problem into the most direct and efficient mathematical form, identifies the required physical attributes, and defines each as a mathematical variable. The second step perceives these attributes from the visual input and assigns a concrete numerical value to each variable. The third step solves the resulting mathematical problem and produces a precise answer.
Natural language is used only in the first step, as interpreting the query and determining its mathematical formulation require linguistic reasoning. The remaining steps operate on numerical values and mathematical expressions, preserving numerical precision and separating perception from computation.

We develop an automated data construction pipeline that generates the required SFT training data for each reasoning step. The model is then trained in multiple stages, with each stage targeting the capability of one reasoning step. Following SFT, GRPO training is applied to improve generalization, with general reasoning problems incorporated to teach the model to distinguish spatial reasoning from other tasks, substantially mitigating the degradation in general reasoning capabilities. Extensive experiments demonstrate that MASIR achieves consistent and substantial improvements across diverse benchmarks, attaining state-of-the-art performance among open-source models across all benchmarks, and even surpassing closed-source models on half of the benchmarks.

The main contributions are as follows:
  \begin{itemize}
  \item We propose MASIR, a new spatial reasoning paradigm that abstracts spatial reasoning problems
  into their inherent mathematical form, mitigating the limitations introduced by natural language
  reasoning.
  \item We develop an automated data construction pipeline and a staged SFT training scheme targeting each reasoning step, followed by GRPO training with general reasoning problems incorporated to preserve broader capabilities.
  \item Ablation studies validate that the performance gains primarily stem from mathematical
  abstraction, confirming the effectiveness of MASIR as a new reasoning paradigm.
  \end{itemize}

\section{Related Work}

\subsection{Multimodal Large Language Models}

Modern MLLMs couple a pre-trained vision encoder to a large language model via a lightweight connector~\cite{llava, internvl, qwen25vltechnicalreport}. To enhance reasoning capabilities, chain-of-thought prompting~\cite{CoT} elicits step-by-step reasoning at inference time, supervised fine-tuning internalizes structured problem-solving behaviors into model weights, and GRPO~\cite{GRPO} optimizes models directly for outcome correctness. However, spatial reasoning still remains a persistent weakness.

\subsection{Spatial Reasoning in MLLMs}

A range of benchmarks have been proposed to evaluate spatial reasoning across single-image, multi-view, and video settings~\cite{3DSRBench, OmniSpatial, Ego3D-Bench, MMSI-Bench, VSI-Bench, STI-Bench, VSR, EmbSpatial-Bench, SpatialRGPT, RoboSpatial, All-Angles, OSI-Bench}.

Existing methods fall into three categories. Prompt-based methods decompose spatial queries using 3D priors or textual cognitive maps~\cite{SpatialPIN, Ego3D-Bench, TRACE}. Training-based methods synthesize spatial datasets and fine-tune MLLMs. SpatialVLM~\cite{SpatialVLM} generates large-scale spatial VQA data for metric distance estimation. SpatialLadder~\cite{SpatialLadder}, SpaceR~\cite{SpaceR}, and Spatial-MLLM~\cite{Spatial-MLLM} further incorporate reinforcement learning, structured representations, or specialized encoders~\cite{MetaSpatial, SpatialReasonor, HiSpatial, 3DThinker}. Tool-based methods introduce external 3D information such as point clouds and depth maps~\cite{SpatialBot, SandboxVLM, SSR}.

Almost all existing methods conduct reasoning in natural language, which loses numerical precision and entangles perception with reasoning. Tool-based methods further require external systems at inference time. MASIR addresses both limitations by abstracting spatial problems into mathematical form and solving them within a single end-to-end model.

\begin{figure*}[htbp]
  \centering
  \includegraphics[width=0.92\textwidth]{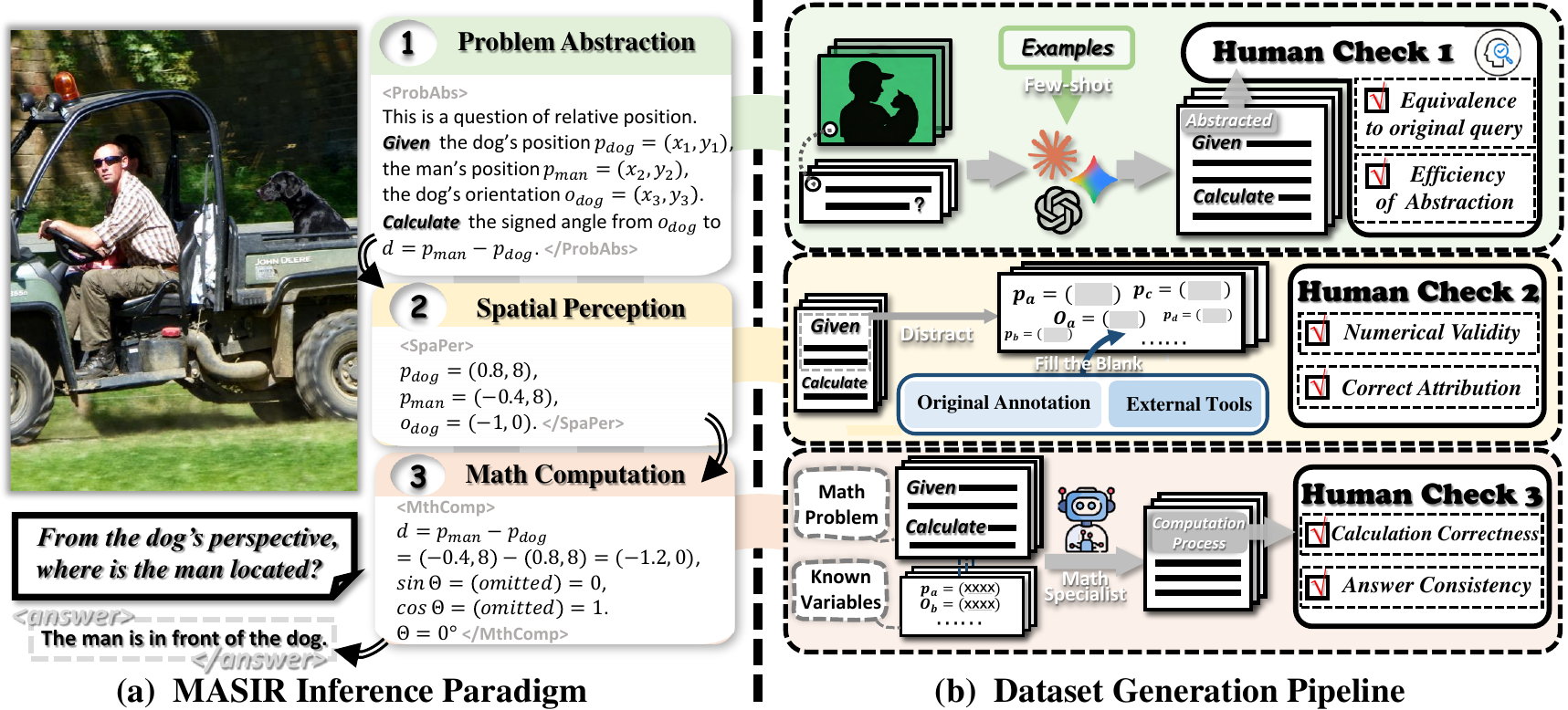}
  \caption{Overview of MASIR inference and data construction. The framework separates problem abstraction, spatial perception, and mathematical computation, with training data built for each step.}
  \label{fig:p_method}
\end{figure*}

\section{MASIR}
\subsection{Overview}
MASIR takes an image $\mathcal{I}$ and a spatial query $q$ as input, and produces an answer $a$ by reformulating the problem into an equivalent mathematical problem. As shown in Figure~\ref{fig:p_method}, the inference is decomposed into three steps: Problem Abstraction, Spatial Perception, and Mathematical Computation, whose outputs are denoted as $T_{\text{ProbAbs}}$, $T_{\text{SpaPer}}$, and $T_{\text{MthComp}}$, respectively.

\subsection{MASIR Inference Paradigm}
\label{sec:method-infer-paradigm}

\begin{figure}[t]
  \centering
  \includegraphics[width=\linewidth]{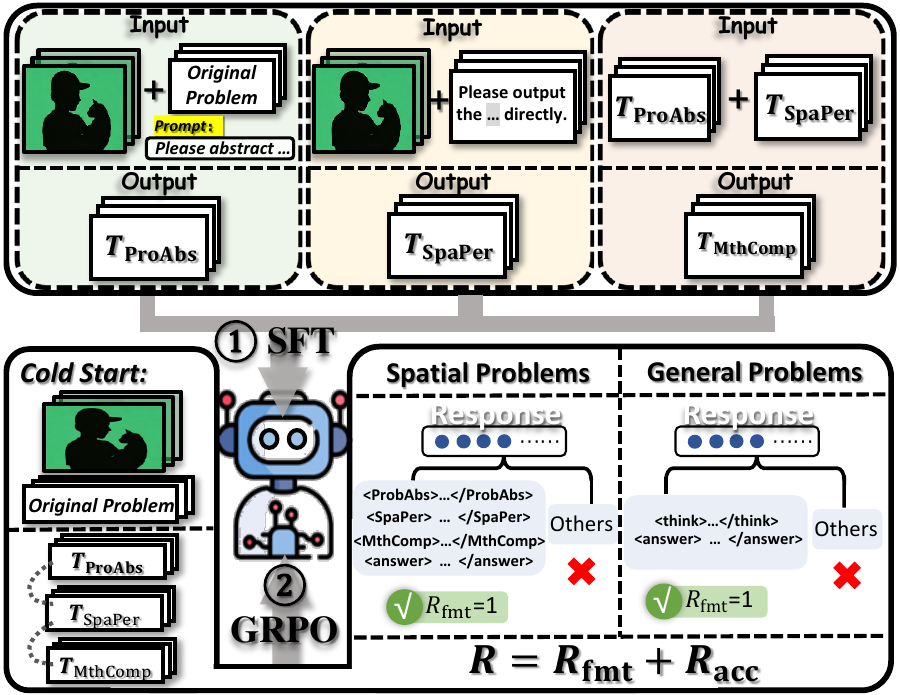}
  \caption{Training pipeline of MASIR.}
  \label{fig:train}
\end{figure}

\subsubsection{Problem Abstraction.}
Given $q$ and $\mathcal{I}$, the first step abstracts $q$ into an equivalent mathematical problem $T_{\text{ProbAbs}}$, with the required physical attributes defined as variables. This step is the core of the entire reasoning process: the formulation produced here determines the structure and efficiency of all subsequent computation. The abstraction identifies the most direct and efficient route to answering $q$, without introducing unnecessary variables or intermediate steps. Interpreting the query and designing its mathematical formulation require linguistic reasoning, making this the only step in MASIR that operates in natural language.

\subsubsection{Spatial Perception.}
Prior work has shown that verbose reasoning chains degrade the perception of MLLMs~\cite{Visutal_degrade}. Therefore, the second step directly outputs the numeric values of the variables defined in $T_{\text{ProbAbs}}$, without any intermediate reasoning. The model reads $\mathcal{I}$ and assigns a concrete value to each variable, producing $T_{\text{SpaPer}}$. All spatial variables are expressed in a camera-centric coordinate system with the camera at the origin. Positions are in meters and orientations are represented as unit vectors. For multi-view inputs, the origin is set to the first view's camera position; for video inputs, to the first frame's camera position.

\subsubsection{Mathematical Computation.}
With $T_{\text{ProbAbs}}$ and $T_{\text{SpaPer}}$ in place, the third step generates $T_{\text{MthComp}}$ by solving the mathematical problem, from which the final answer $a$ is derived. Since the abstracted problems are elementary, current MLLMs can solve them with high reliability.

\subsection{Data Construction Pipeline}
\label{sec:method-data-pipeline}
The data construction pipeline follows the same three-step structure as the inference paradigm (\S~\ref{sec:method-infer-paradigm}). The source data consists of 20.7k image-text pairs from the spatial training subset of SpatialLadder~\cite{SpatialLadder}, covering 7 question types: relative direction, relative and absolute distance, object size, counting, room size, and appearance order. Training data for each step is constructed independently.

\subsubsection{Problem Abstraction.}
For each of the 7 question types, five representative questions are manually abstracted into mathematical form as few-shot examples. These examples are provided to frontier models\cite{anthropic2026claudeopus47,openai2026gpt55,google2026gemini31pro} to generate mathematical abstractions for all 20.7k samples. All generated samples are manually verified for equivalence to the original query and efficiency of abstraction, where efficiency requires that the abstracted problem introduces no redundant variables and reaches the answer through the most direct computational path. More details and generation prompts are in Appendix A.1.

\subsubsection{Spatial Perception.}
For each abstracted problem in $T_{\text{ProbAbs}}$, the required variable values are obtained from two sources. For coordinate-type variables, the original metadata is traced back through SpatialLadder's annotations to the source 3D dataset\cite{dai2017scannet}, from which the 3D bounding boxes are retrieved and processed into the required coordinates. For other attributes such as orientation and depth, external tools\cite{yang2024depthanything,wang2024orientanything} are applied directly to the images. All assigned values are manually verified for numerical validity and correct attribution. Correct attribution means each value is confirmed to correspond to the correct object and attribute as defined in $T_{\text{ProbAbs}}$, ensuring no mismatch between the assigned values and the intended variables. Details of coordinate transformation and the calibration procedure for externally estimated attributes are provided in Appendix A.2.

\subsubsection{Mathematical Computation.}
For each sample, DeepSeek-Math~\cite{GRPO} generates the computation $T_{\text{MthComp}}$ from $T_{\text{ProbAbs}}$ and $T_{\text{SpaPer}}$. All generated samples are manually verified on two criteria: the correctness of each computation step, and the consistency of the final answer with the ground truth.
\subsection{Training}
\label{sec:method-training}
\subsubsection{Supervised Fine-Tuning.}
Three types of training data are constructed to develop each capability independently, and the model is trained on all three simultaneously.

\begin{itemize}
\item \textbf{Dataset for Problem Abstraction.} 

\textit{Input}: image $\mathcal{I}$, query $q$, and the prompt \textit{``Please abstract the spatial reasoning question into its mathematical form.''}
\textit{Output}: $T_{\text{ProbAbs}}$.

\item \textbf{Dataset for Spatial Perception.} 

\textit{Input}: image $\mathcal{I}$ and a perception prompt derived from the variable names in $T_{\text{ProbAbs}}$ (e.g., \textit{``Please output the position of the dog in the picture directly.''} for ``$p_{dog}$''). \textit{Output}: $T_{\text{SpaPer}}$.

\item \textbf{Dataset for Mathematical Computation.} 

\textit{Input}: $T_{\text{ProbAbs}}$ and $T_{\text{SpaPer}}$. \textit{Output}: $T_{\text{MthComp}}$.
\end{itemize}

\subsubsection{GRPO.}
A cold start SFT is performed first using the 1,255 high-quality samples selected by SpatialLadder~\cite{SpatialLadder}. For each sample, $T_{\text{ProbAbs}}$, $T_{\text{SpaPer}}$, and $T_{\text{MthComp}}$ are concatenated into a complete reasoning chain. This ensures the model can reliably generate outputs that conform to the required format before GRPO training.

GRPO~\cite{GRPO} is then applied to further improve generalization. Since the MASIR reasoning paradigm is designed specifically for spatial reasoning problems, applying it to general visual reasoning problems may introduce format bias into the responses, degrading the model's general reasoning capabilities. To prevent this, ThinkLite-VL-hard-11k~\cite{thinklite} is incorporated into the GRPO training data, forming $\mathcal{D}_{\text{spatial}}$ and $\mathcal{D}_{\text{general}}$. Two format rewards are defined for the two data sources:
\begin{equation}
R_{\text{fmt}}(q, \hat{y}) = \begin{cases}
1 & \text{if } \hat{y} \text{ matches \textit{Form 1} and } q \in \mathcal{D}_{\text{spatial}}, \\
1 & \text{if } \hat{y} \text{ matches \textit{Form 2} and } q \in \mathcal{D}_{\text{general}}, \\
0 & \text{otherwise}.
\end{cases}
\end{equation}
Here, $\hat{y}$ denotes the model's generated response and $q$ denotes the input question. \textit{Form 1} requires MASIR reasoning tags \texttt{<ProbAbs>}, \texttt{<SpaPer>}, \texttt{<MthComp>} and \texttt{<answer>} to be present and well-formed, and \textit{Form 2} requires the standard reasoning tags \texttt{<think>} and \texttt{<answer>}. This design encourages the model to distinguish between spatial reasoning problems and general visual reasoning problems, applying different reasoning paradigms accordingly, and prevents over-generalization of the MASIR format from degrading general reasoning capabilities.

The accuracy reward varies based on the question type. For multiple-choice questions, $R_{\text{mc}}$ is defined as:
\begin{equation}
R_{\text{mc}}(\hat{y}, y) = \begin{cases} 1 & \text{if } \hat{y} = y, \\ 0 & \text{otherwise.} \end{cases}
\end{equation}
For numerical questions, $R_{\text{num}}$ follows the Mean Relative Accuracy (MRA) metric introduced by VSI-Bench~\cite{VSI-Bench}, evaluating predictions over a set of relative-error thresholds $\theta_i \in \{0.5, 0.55, \dots, 0.95\}$:
\begin{equation}
R_{\text{num}}(\hat{y}, y) = \frac{1}{N}\sum_{i=1}^{N} \mathbf{1}\left( \frac{|\hat{y} - y|}{y} \leq 1 - \theta_i \right),
\end{equation}
where $\hat{y}$ denotes the model's predicted answer, $y$ denotes the ground truth, and $N$ is the total number of thresholds. All ground-truth values in the training data satisfy $y > 0$. This formulation rewards predictions that fall within progressively tighter tolerance bands, producing a smoother training signal than binary judgment.

The total reward is $R = R_{\text{acc}} + R_{\text{fmt}}$. For each question, a group of responses is sampled from the current policy, and group-relative advantages are computed to update the policy, following the standard GRPO objective~\cite{GRPO}.

\section{Experiments}
\subsection{Experimental Setup}
\subsubsection{Implementation Details.}
All models are trained on 4$\times$A100 GPUs. SFT stage uses a batch size of 4 and runs for 1 epoch. GRPO stage uses a batch size of 1 per device, runs for 2 epochs, and generates 8 responses per sample. All experiments are run with a temperature of 0.01. Each result is averaged over three independent runs to ensure stability.

\subsubsection{Benchmarks.}
MASIR is evaluated on six benchmarks spanning three input modalities.
\begin{itemize}
\item \textbf{Single-image.} 3DSRBench~\cite{3DSRBench} evaluates basic 3D geometric perception of perspective, depth, and scale. OmniSpatial~\cite{OmniSpatial} covers higher-order spatial cognition including dynamic evolution, object interaction, and multi-step logic.
\item \textbf{Multi-view.} Ego3D-Bench~\cite{Ego3D-Bench} focuses on egocentric metric 3D reasoning in autonomous driving scenarios with synchronous surround-camera inputs. MMSI-Bench~\cite{MMSI-Bench} is an indoor multi-view benchmark emphasizing scene layout reconstruction and cross-view geometric correspondence.
\item \textbf{Video.} VSI-Bench~\cite{VSI-Bench} evaluates spatio-temporal memory and spatial layout reasoning in indoor walkthrough videos. STI-Bench~\cite{STI-Bench} evaluates quantitative spatial measurement and dynamic temporal motion reasoning across desktop, indoor and outdoor real-world videos.
\end{itemize}

We verified using the available scene, video, and image identifiers that no evaluation instance overlaps with the SpatialLadder-derived training data.

\subsubsection{Baselines.}
Three categories of baselines are included. 
\begin{itemize}
\item \textbf{Closed-source models.} Claude 3.7 Sonnet~\cite{anthropic2025claude37sonnet}, GPT-4o~\cite{gpt4o}, and Gemini 2.5 Pro~\cite{gemini-25-pro}. 
\item \textbf{Open-source models.} Qwen2.5-VL-3B, Qwen2.5-VL-7B~\cite{qwen25vltechnicalreport}, Qwen3-VL-8B~\cite{qwen3vl}, and InternVL3-8B~\cite{internvl3}; MASIR is applied to all four models to verify its effectiveness across different backbones. 
\item \textbf{Spatial reasoning models.} The spatial models from Spatial-MLLM~\cite{Spatial-MLLM}, SpatialLadder~\cite{SpatialLadder}, 3DThinker~\cite{3DThinker}, SpaceR~\cite{SpaceR}, and VILASR~\cite{VILASR} are also included for comparison.
\end{itemize}

\begin{table*}[t]
\centering
\small
\begin{tabular}{lllllll}
\toprule
\multirow{2}{*}{Model} & \multicolumn{2}{c}{\textbf{Single-image}} & \multicolumn{2}{c}{\textbf{Multi-View}} & \multicolumn{2}{c}{\textbf{Video}} \\
\cmidrule(lr){2-3} \cmidrule(lr){4-5} \cmidrule(lr){6-7}
& 3DSRBench & OmniSpatial & Ego3D-Bench & MMSI-Bench & VSI-Bench & STI-Bench \\
\midrule
\multicolumn{7}{c}{\textit{Closed-source Models}} \\
Claude 3.7 Sonnet & 48.27 & 48.91 & 53.92 & 30.35 & 48.81 & 40.56 \\
GPT-4o & 44.21 & 47.81 & 48.18 & 30.39 & 34.02 & 34.83 \\
Gemini 2.5 Pro & 48.47 & 55.19 & 66.75 & 36.94 & 51.53 & 41.49 \\
\midrule
\multicolumn{7}{c}{\textit{Qwen2.5-VL-3B based Spatial Models}} \\
Qwen2.5-VL-3B & 43.95 & 40.36 & 39.17 & 26.51 & 29.48 & 32.13 \\
SpatialLadder-3B & 44.56\textcolor{diffgreen}{\scriptsize{(+0.61)}} & 41.15\textcolor{diffgreen}{\scriptsize{(+0.79)}} & 42.04\textcolor{diffgreen}{\scriptsize{(+2.87)}} & 29.26\textcolor{diffgreen}{\scriptsize{(+2.75)}} & 45.79\textcolor{diffgreen}{\scriptsize{(+16.31)}} & 31.94\textcolor{diffred}{\scriptsize{(-0.19)}} \\
Spatial-MLLM-4B & \underline{45.74}\textcolor{diffgreen}{\scriptsize{(+1.79)}} & \underline{43.74}\textcolor{diffgreen}{\scriptsize{(+3.38)}} & \underline{44.78}\textcolor{diffgreen}{\scriptsize{(+5.61)}} & \underline{31.57}\textcolor{diffgreen}{\scriptsize{(+5.06)}} & \underline{47.36}\textcolor{diffgreen}{\scriptsize{(+17.88)}} & \underline{33.94}\textcolor{diffgreen}{\scriptsize{(+1.81)}} \\
\rowcolor{gray!15} MASIR $_{\text{\tiny Qwen2.5-VL-3B}}$ & \textbf{47.08}\textcolor{diffgreen}{\scriptsize{(+3.13)}} & \textbf{48.14}\textcolor{diffgreen}{\scriptsize{(+7.78)}} & \textbf{52.27}\textcolor{diffgreen}{\scriptsize{(+13.10)}} & \textbf{38.71}\textcolor{diffgreen}{\scriptsize{(+12.20)}} & \textbf{48.71}\textcolor{diffgreen}{\scriptsize{(+19.23)}} & \textbf{37.79}\textcolor{diffgreen}{\scriptsize{(+5.66)}} \\
\midrule
\multicolumn{7}{c}{\textit{Qwen2.5-VL-7B based Spatial Models}} \\
Qwen2.5-VL-7B & 48.43 & 39.18 & 41.12 & 25.98 & 35.87 & 34.51 \\
VILASR-7B & 48.62\textcolor{diffgreen}{\scriptsize{(+0.19)}} & \underline{42.06}\textcolor{diffgreen}{\scriptsize{(+2.88)}} & \underline{46.75}\textcolor{diffgreen}{\scriptsize{(+5.63)}} & \underline{32.24}\textcolor{diffgreen}{\scriptsize{(+6.26)}} & \underline{45.41}\textcolor{diffgreen}{\scriptsize{(+9.54)}} & 36.18\textcolor{diffgreen}{\scriptsize{(+1.67)}} \\
SpaceR-7B & \underline{49.84}\textcolor{diffgreen}{\scriptsize{(+1.41)}} & 42.20\textcolor{diffgreen}{\scriptsize{(+3.02)}} & 45.67\textcolor{diffgreen}{\scriptsize{(+4.55)}} & 29.83\textcolor{diffgreen}{\scriptsize{(+3.85)}} & 44.52\textcolor{diffgreen}{\scriptsize{(+8.65)}} & \underline{37.08}\textcolor{diffgreen}{\scriptsize{(+2.57)}} \\
\rowcolor{gray!15} MASIR $_{\text{\tiny Qwen2.5-VL-7B}}$ & \textbf{52.96}\textcolor{diffgreen}{\scriptsize{(+4.53)}} & \textbf{49.17}\textcolor{diffgreen}{\scriptsize{(+9.99)}} & \textbf{54.91}\textcolor{diffgreen}{\scriptsize{(+13.79)}} & \textbf{39.11}\textcolor{diffgreen}{\scriptsize{(+13.13)}} & \textbf{52.84}\textcolor{diffgreen}{\scriptsize{(+16.97)}} & \textbf{40.06}\textcolor{diffgreen}{\scriptsize{(+5.55)}} \\
\midrule
\multicolumn{7}{c}{\textit{Other Open-source Models}} \\
Qwen3-VL-8B & 52.84 & 46.11 & 42.51 & 26.94 & 59.47 & 31.25 \\
\rowcolor{gray!15} MASIR $_{\text{\tiny Qwen3-VL-8B}}$ & {54.77}\textcolor{diffgreen}{\scriptsize{(+1.93)}} & {49.51}\textcolor{diffgreen}{\scriptsize{(+3.40)}} & {55.06}\textcolor{diffgreen}{\scriptsize{(+12.55)}} & {37.49}\textcolor{diffgreen}{\scriptsize{(+10.55)}} & {61.88}\textcolor{diffgreen}{\scriptsize{(+2.41)}} & {34.91}\textcolor{diffgreen}{\scriptsize{(+3.66)}} \\
InternVL3-8B & 50.51 & 41.63 & 43.19 & 25.72 & 38.15 & 30.78 \\
\rowcolor{gray!15} MASIR $_{\text{\tiny InternVL3-8B}}$ & {57.42}\textcolor{diffgreen}{\scriptsize{(+6.91)}} & {46.17}\textcolor{diffgreen}{\scriptsize{(+4.54)}} & {50.74}\textcolor{diffgreen}{\scriptsize{(+7.55)}} & {33.28}\textcolor{diffgreen}{\scriptsize{(+7.56)}} & {41.84}\textcolor{diffgreen}{\scriptsize{(+3.69)}} & {33.74}\textcolor{diffgreen}{\scriptsize{(+2.96)}} \\
\bottomrule
\end{tabular}
\caption{Main results on six spatial reasoning benchmarks. Green values indicate improvements and red values indicate decrements compared to the corresponding base model.}
\label{tab:main_results}
\end{table*}

\subsection{Main Results}
To verify the effectiveness of MASIR as a spatial reasoning paradigm, it is applied to four base models and evaluated against closed-source models, open-source base models, and existing spatial reasoning models across three benchmark categories: single-image, multi-view, and video.

Results are reported in Table~\ref{tab:main_results}. MASIR consistently improves over all four base models, with gains observed across all three benchmark categories. This holds regardless of model size and architecture, indicating that the mathematical abstraction paradigm generalizes across different backbones.

Among models built on Qwen2.5-VL-3B, MASIR outperforms SpatialLadder and Spatial-MLLM across every benchmark. Among models built on Qwen2.5-VL-7B, MASIR surpasses SpaceR and VILASR on all six benchmarks, with particularly large margins on Ego3D-Bench at 54.91 compared to 46.75 for VILASR and 45.67 for SpaceR.

MASIR applied to Qwen2.5-VL-7B surpasses Claude 3.7 Sonnet on 3DSRBench at 52.96 versus 48.27, and on VSI-Bench at 52.84 versus 48.81. It also exceeds GPT-4o on all six benchmarks. These results show that the MASIR paradigm enables open-source models to reach competitive performance with proprietary systems on spatial reasoning tasks.

\subsection{Math Representation Drives the Gains}
To further verify that the performance gains stem from the mathematical representation rather than the training structure, a controlled natural language variant, NL-MASIR, is constructed to match MASIR in all aspects except the representation format of intermediate reasoning.

NL-MASIR uses the same images, questions, spatial attributes, numerical values, computation order, and final answers as MASIR, with intermediate reasoning expressed in natural language. The natural language data is generated using fixed templates to avoid data quality differences. The conversion templates are provided in the Appendix B.

Both variants are trained with identical configurations, including training samples, SFT settings, GRPO settings, reward functions, and training steps. The sole controlled variable is the representation format.

\begin{table}[t]
  \centering
  \footnotesize
  \setlength{\tabcolsep}{4pt}
  \begin{tabular}{llll}
  \toprule
  Model & 3DSRBench & MMSI-Bench & VSI-Bench \\
  \midrule
  \multicolumn{4}{c}{\textit{Qwen2.5-VL-3B}} \\
  base   & 43.95 & 26.51 & 29.48 \\
  NL-MASIR & 44.29\rlap{\textcolor{diffgreen}{\scriptsize{(+0.34)}}} & 29.16\rlap{\textcolor{diffgreen}{\scriptsize{(+2.65)}}} & 33.91\rlap{\textcolor{diffgreen}{\scriptsize{(+4.43)}}} \\
  MASIR & 47.08\rlap{\textcolor{diffgreen}{\scriptsize{(+3.13)}}} & 38.71\rlap{\textcolor{diffgreen}{\scriptsize{(+12.20)}}} & 48.71\rlap{\textcolor{diffgreen}{\scriptsize{(+19.23)}}} \\
  \midrule
  \multicolumn{4}{c}{\textit{Qwen3-VL-8B}} \\
  base   & 52.84 & 26.94 & 59.47 \\
  NL-MASIR & 53.13\rlap{\textcolor{diffgreen}{\scriptsize{(+0.29)}}} & 28.96\rlap{\textcolor{diffgreen}{\scriptsize{(+2.02)}}} & 59.59\rlap{\textcolor{diffgreen}{\scriptsize{(+0.12)}}} \\
  MASIR & 54.77\rlap{\textcolor{diffgreen}{\scriptsize{(+1.93)}}} & 37.49\rlap{\textcolor{diffgreen}{\scriptsize{(+10.55)}}} & 61.88\rlap{\textcolor{diffgreen}{\scriptsize{(+2.41)}}} \\
  \midrule
  \multicolumn{4}{c}{\textit{InternVL3-8B}} \\
  base   & 50.51 & 25.72 & 38.15 \\
  NL-MASIR & 51.82\rlap{\textcolor{diffgreen}{\scriptsize{(+1.31)}}} & 26.91\rlap{\textcolor{diffgreen}{\scriptsize{(+1.19)}}} & 39.01\rlap{\textcolor{diffgreen}{\scriptsize{(+0.86)}}} \\
  MASIR & 57.42\rlap{\textcolor{diffgreen}{\scriptsize{(+6.91)}}} & 33.28\rlap{\textcolor{diffgreen}{\scriptsize{(+7.56)}}} & 41.84\rlap{\textcolor{diffgreen}{\scriptsize{(+3.69)}}} \\
  \bottomrule
  \end{tabular}
  \caption{Ablation on natural language reasoning variant NL-MASIR and mathematical MASIR.}
  \label{tab:ablation_nl_masir}
\end{table}
\begin{table}[t]
\centering
\small
\begin{tabular}{llll}
\toprule
Model & 3DSRBench & MMSI-Bench & VSI-Bench \\
\midrule
\multicolumn{4}{c}{\textit{Qwen2.5-VL-3B}} \\
MASIR           & 47.08 & 38.71 & 48.71 \\
\textit{w/o} SFT on T1 & 44.67\rlap{\scriptsize\textcolor{diffred}{(-2.41)}} & 30.18\rlap{\scriptsize\textcolor{diffred}{(-8.53)}} & 37.19\rlap{\scriptsize\textcolor{diffred}{(-11.52)}} \\
\textit{w/o} SFT on T2  & 46.05\rlap{\scriptsize\textcolor{diffred}{(-1.03)}} & 32.31\rlap{\scriptsize\textcolor{diffred}{(-6.40)}} & 40.54\rlap{\scriptsize\textcolor{diffred}{(-8.17)}} \\
\textit{w/o} SFT on T3 & 46.48\rlap{\scriptsize\textcolor{diffred}{(-0.60)}} & 35.14\rlap{\scriptsize\textcolor{diffred}{(-3.57)}} & 42.35\rlap{\scriptsize\textcolor{diffred}{(-6.36)}} \\
\midrule
\multicolumn{4}{c}{\textit{Qwen3-VL-8B}} \\
MASIR           & 54.77 & 37.49 & 61.88 \\
\textit{w/o} SFT on T1 & 53.48\rlap{\scriptsize\textcolor{diffred}{(-1.29)}} & 30.41\rlap{\scriptsize\textcolor{diffred}{(-7.08)}} & 60.51\rlap{\scriptsize\textcolor{diffred}{(-1.37)}} \\
\textit{w/o} SFT on T2  & 53.96\rlap{\scriptsize\textcolor{diffred}{(-0.81)}} & 34.90\rlap{\scriptsize\textcolor{diffred}{(-2.59)}} & 60.92\rlap{\scriptsize\textcolor{diffred}{(-0.96)}} \\
\textit{w/o} SFT on T3 & 54.19\rlap{\scriptsize\textcolor{diffred}{(-0.58)}} & 34.67\rlap{\scriptsize\textcolor{diffred}{(-2.82)}} & 61.40\rlap{\scriptsize\textcolor{diffred}{(-0.48)}} \\
\midrule
\multicolumn{4}{c}{\textit{InternVL3-8B}} \\
MASIR           & 57.42 & 33.28 & 41.84 \\
\textit{w/o} SFT on T1 & 53.16\rlap{\scriptsize\textcolor{diffred}{(-4.26)}} & 28.10\rlap{\scriptsize\textcolor{diffred}{(-5.18)}} & 39.30\rlap{\scriptsize\textcolor{diffred}{(-2.54)}} \\
\textit{w/o} SFT on T2  & 54.63\rlap{\scriptsize\textcolor{diffred}{(-2.79)}} & 30.02\rlap{\scriptsize\textcolor{diffred}{(-3.26)}} & 39.91\rlap{\scriptsize\textcolor{diffred}{(-1.93)}} \\
\textit{w/o} SFT on T3 & 54.61\rlap{\scriptsize\textcolor{diffred}{(-2.81)}} & 31.30\rlap{\scriptsize\textcolor{diffred}{(-1.98)}} & 41.00\rlap{\scriptsize\textcolor{diffred}{(-0.84)}} \\
\bottomrule
\end{tabular}
\caption{Ablation study on the contribution of each reasoning step. T1, T2, and T3 denote $T_{\text{ProbAbs}}$, $T_{\text{SpaPer}}$, and $T_{\text{MthComp}}$.}
\label{tab:ablation_step_sft}
\end{table}

Table~\ref{tab:ablation_nl_masir} reports results for Base, NL-MASIR, and MASIR. The gap between Base and NL-MASIR reflects the effect of structured supervision, and the gap between NL-MASIR and MASIR reflects the additional contribution of mathematical representation. MASIR outperforms NL-MASIR across all benchmarks, providing evidence that the gains stem from the mathematical abstraction paradigm rather than the training structure alone.

\subsection{Contribution of Each Reasoning Step}
To examine the contribution of each reasoning step to MASIR's performance, ablation experiments are conducted by removing the training data for each step independently while keeping the remaining two intact. In each ablated variant, the model still generates all three reasoning steps during inference, but the removed step receives no dedicated training. All other training details, including details of GRPO, remain identical across variants.

As shown in Table~\ref{tab:ablation_step_sft}, removing $T_{\text{ProbAbs}}$ SFT leads to the most significant performance degradation across all base models, with accuracy drops exceeding 8 points on the Ego3D-Bench and MMSI-Bench. This observation demonstrates that problem abstraction constitutes the most critical component of the MASIR paradigm, verifying our core premise that mathematical abstraction serves as the primary driver for boosting spatial reasoning performance.

Removing the $T_{\text{SpaPer}}$ SFT stage also results in consistent performance declines across all benchmarks, highlighting the essential role of dedicated spatial perception in spatial reasoning tasks. Without targeted spatial perception training, the model struggles to precisely assign physical spatial values to the abstract variables defined in $T_{\text{ProbAbs}}$, thereby inherently limiting the correctness of subsequent computational reasoning.

In contrast, ablating $T_{\text{MthComp}}$ SFT yields the smallest performance drops. This phenomenon is reasonable, as modern MLLMs already inherit powerful inherent mathematical computation capabilities. Even without specific supervision, the model can reliably complete most computational operations, leading to the marginal and least sensitive contribution of the mathematical computation module among the three reasoning steps.

\subsection{Contribution of Training Components}
To verify the necessity of each training stage, ablations are conducted by applying SFT and GRPO independently. Three typical benchmarks covering single-image, multi-view, and video reasoning are selected, along with three base models of different sizes and architectures, to ensure the consistency of the conclusions. Results are reported in Table~\ref{tab:ablation_training}.

Both SFT and GRPO contribute to the final performance. Applying either stage alone improves over the base model.
Among the two stages, GRPO alone yields larger gains than SFT alone on most benchmarks. This is because SFT trains each reasoning step independently without optimizing the coordination among them, whereas GRPO trains on complete reasoning chains and directly optimizes how the three steps work together to produce the final answer. Together, the two stages are complementary, and removing either one reduces overall performance.

\begin{table}[t]
  \centering
  \footnotesize
  \setlength{\tabcolsep}{4pt}
  \begin{tabular}{llll}
  \toprule
  Model & 3DSRBench & MMSI-Bench & VSI-Bench \\
  \midrule
  \multicolumn{4}{c}{\textit{Qwen2.5-VL-3B}} \\
  base   & 43.95 & 26.51 & 29.48 \\
  + SFT only & 45.13\rlap{\textcolor{diffgreen}{\scriptsize{(+1.18)}}} & 30.10\rlap{\textcolor{diffgreen}{\scriptsize{(+3.59)}}} & 33.19\rlap{\textcolor{diffgreen}{\scriptsize{(+3.71)}}} \\
  + GRPO only & 46.41\rlap{\textcolor{diffgreen}{\scriptsize{(+2.46)}}} & 31.09\rlap{\textcolor{diffgreen}{\scriptsize{(+4.58)}}} & 36.98\rlap{\textcolor{diffgreen}{\scriptsize{(+7.50)}}} \\
  MASIR & 47.08\rlap{\textcolor{diffgreen}{\scriptsize{(+3.13)}}} & 38.71\rlap{\textcolor{diffgreen}{\scriptsize{(+12.20)}}} & 48.71\rlap{\textcolor{diffgreen}{\scriptsize{(+19.23)}}} \\
  \midrule
  \multicolumn{4}{c}{\textit{Qwen3-VL-8B}} \\
  base   & 52.84 & 26.94 & 59.47 \\
  + SFT only & 53.61\rlap{\textcolor{diffgreen}{\scriptsize{(+0.77)}}} & 29.03\rlap{\textcolor{diffgreen}{\scriptsize{(+2.09)}}} & 60.05\rlap{\textcolor{diffgreen}{\scriptsize{(+0.58)}}} \\
  + GRPO only & 53.91\rlap{\textcolor{diffgreen}{\scriptsize{(+1.07)}}} & 31.96\rlap{\textcolor{diffgreen}{\scriptsize{(+5.02)}}} & 60.90\rlap{\textcolor{diffgreen}{\scriptsize{(+1.43)}}} \\
  MASIR & 54.77\rlap{\textcolor{diffgreen}{\scriptsize{(+1.93)}}} & 37.49\rlap{\textcolor{diffgreen}{\scriptsize{(+10.55)}}} & 61.88\rlap{\textcolor{diffgreen}{\scriptsize{(+2.41)}}} \\
  \midrule
  \multicolumn{4}{c}{\textit{InternVL3-8B}} \\
  base   & 50.51 & 25.72 & 38.15 \\
  + SFT only & 53.90\rlap{\textcolor{diffgreen}{\scriptsize{(+3.39)}}} & 27.30\rlap{\textcolor{diffgreen}{\scriptsize{(+1.58)}}} & 39.60\rlap{\textcolor{diffgreen}{\scriptsize{(+1.45)}}} \\
  + GRPO only & 55.54\rlap{\textcolor{diffgreen}{\scriptsize{(+5.03)}}} & 29.90\rlap{\textcolor{diffgreen}{\scriptsize{(+4.18)}}} & 40.93\rlap{\textcolor{diffgreen}{\scriptsize{(+2.78)}}} \\
  MASIR & 57.42\rlap{\textcolor{diffgreen}{\scriptsize{(+6.91)}}} & 33.28\rlap{\textcolor{diffgreen}{\scriptsize{(+7.56)}}} & 41.84\rlap{\textcolor{diffgreen}{\scriptsize{(+3.69)}}} \\
  \bottomrule
  \end{tabular}
  \caption{Effect of training components. MASIR denotes the full training pipeline with both SFT and GRPO applied. `+GRPO only' includes the cold-start SFT stage.}
  \label{tab:ablation_training}
\end{table}

\begin{table}[t]
  \centering
  \footnotesize
  \setlength{\tabcolsep}{4pt}
  \begin{tabular*}{\linewidth}{@{\extracolsep{\fill}} llll @{}}
  \toprule
  Model & MMStar & MMMU & VisuLogic \\
  \midrule
  \multicolumn{4}{c}{\textit{Qwen2.5-VL-3B}} \\
  Base                              & 55.90 & 53.10 & 24.90 \\
  MASIR w/o $\mathcal{D}_{\text{general}}$ & 44.12\rlap{\textcolor{diffred}{\scriptsize{(-11.78)}}} & 39.14\rlap{\textcolor{diffred}{\scriptsize{(-13.96)}}} & 18.37\rlap{\textcolor{diffred}{\scriptsize{(-6.53)}}} \\
  MASIR w/ $\mathcal{D}_{\text{general}}$  & 54.92\rlap{\textcolor{diffred}{\scriptsize{(-0.98)}}} & 53.07\rlap{\textcolor{diffred}{\scriptsize{(-0.03)}}} & 24.56\rlap{\textcolor{diffred}{\scriptsize{(-0.34)}}} \\
  \midrule
  \multicolumn{4}{c}{\textit{Qwen3-VL-8B}} \\
  Base                              & 70.90 & 55.90 & 22.50 \\
  MASIR w/o $\mathcal{D}_{\text{general}}$ & 56.70\rlap{\textcolor{diffred}{\scriptsize{(-14.20)}}} & 48.10\rlap{\textcolor{diffred}{\scriptsize{(-7.80)}}} & 16.90\rlap{\textcolor{diffred}{\scriptsize{(-5.60)}}} \\
  MASIR w/ $\mathcal{D}_{\text{general}}$  & 70.10\rlap{\textcolor{diffred}{\scriptsize{(-0.80)}}} & 55.16\rlap{\textcolor{diffred}{\scriptsize{(-0.74)}}} & 22.08\rlap{\textcolor{diffred}{\scriptsize{(-0.42)}}} \\
  \midrule
  \multicolumn{4}{c}{\textit{InternVL3-8B}} \\
  Base                              & 68.20 & 62.70 & 25.90 \\
  MASIR w/o $\mathcal{D}_{\text{general}}$ & 57.80\rlap{\textcolor{diffred}{\scriptsize{(-10.40)}}} & 55.02\rlap{\textcolor{diffred}{\scriptsize{(-7.68)}}} & 20.40\rlap{\textcolor{diffred}{\scriptsize{(-5.50)}}} \\
  MASIR w/ $\mathcal{D}_{\text{general}}$  & 67.56\rlap{\textcolor{diffred}{\scriptsize{(-0.64)}}} & 62.42\rlap{\textcolor{diffred}{\scriptsize{(-0.28)}}} & 25.67\rlap{\textcolor{diffred}{\scriptsize{(-0.23)}}} \\
  \bottomrule
  \end{tabular*}
  \caption{Effect on general reasoning capability. MASIR w/ $\mathcal{D}_{\text{general}}$ denotes
  the full MASIR training with ThinkLite-VL-hard-11k included in GRPO, and w/o
  $\mathcal{D}_{\text{general}}$ denotes its removal.}
  \label{tab:ablation_general}
\end{table}

\subsection{Effect on General Capability}
To examine whether MASIR training degrades general reasoning capability, a variant is constructed by removing $\mathcal{D}_{\text{general}}$ from the GRPO stage, leaving all other training details unchanged. This variant is evaluated on three general reasoning benchmarks across three base models.

As shown in Table~\ref{tab:ablation_general}, without $\mathcal{D}_{\text{general}}$, performance drops by over 10 points on MMStar across all three base models. The model tends to apply MASIR-specific reasoning tags to general problems, disrupting the normal reasoning process and producing erroneous outputs.

Including $\mathcal{D}_{\text{general}}$ reduces the drops to under 1 point across all models and benchmarks. The model learns to distinguish between spatial reasoning problems and general visual reasoning problems, applying different reasoning paradigms accordingly. This confirms that the proposed training design effectively preserves general reasoning capability.

\subsection{Further Analysis}
\subsubsection{Where Does MASIR Improve Most?}
\begin{figure}[t]
  \centering
  \includegraphics[width=0.92\linewidth]{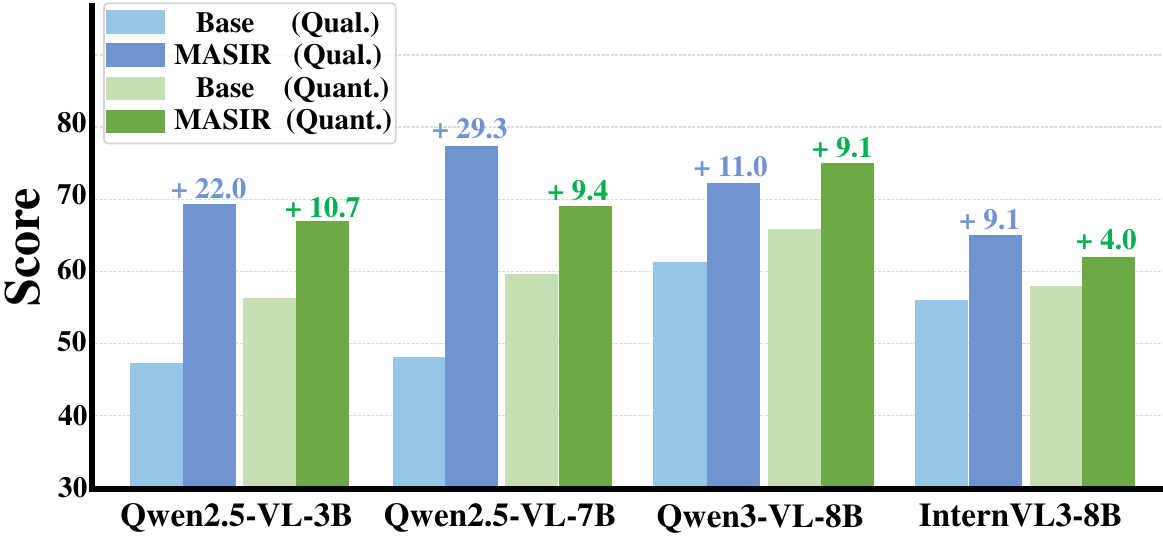}
  \caption{Performance comparison on qualitative and quantitative visual reasoning tasks across four base models.}
  \label{fig:analysis1}
\end{figure}

Spatial reasoning problems can be categorized into two types: qualitative and quantitative.
Qualitative problems require discrete categorical answers, such as direction judgment. Quantitative problems require continuous numerical answers, such as absolute distance and object size. To analyze which type benefits more from MASIR, all sub-tasks across the six benchmarks are grouped into these two categories, and the average improvement of MASIR over the base model is computed for each. Results are shown in Figure~\ref{fig:analysis1}.

MASIR consistently achieves larger improvements on qualitative problems. These problems require complex spatial relationship judgments in natural language space; mathematical abstraction removes this burden by enabling more direct computation. For quantitative problems, the bottleneck lies in perceptual accuracy: errors in perceived numerical values propagate to the final answer even when the abstraction is correct. Since MASIR does not directly improve perception, gains on this category are relatively limited.

\subsubsection{Error Analysis.}
\begin{figure}[t]
  \centering
  \includegraphics[width=0.98\linewidth]{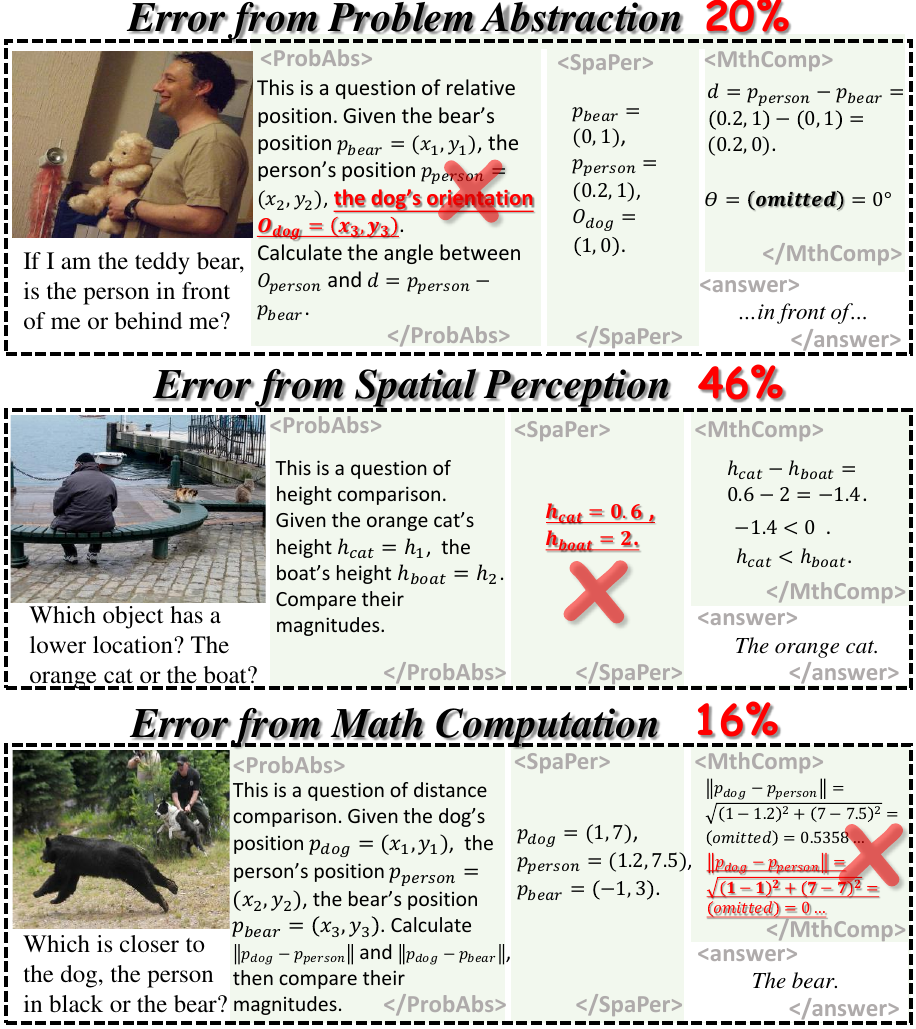}
  \caption{Error analysis of MASIR failure cases.}
  \label{fig:analysis2}
\end{figure}

To understand where MASIR fails, all incorrect predictions are examined and labeled by human annotators. Each failure is attributed to the first step where an error occurs; samples with errors in multiple steps are counted only once. Figure~\ref{fig:analysis2} shows representative examples of each error type.

Spatial Perception accounts for 46\% of failures, indicating that the primary bottleneck lies in accurately reading physical attributes from visual input, not in the reasoning structure itself. Problem Abstraction contributes 20\%, showing that MLLMs can largely learn to correctly abstract spatial queries into mathematical form. Mathematical Computation errors are the least frequent at 16\%, consistent with the observation that the abstracted problems are typically elementary for modern MLLMs. The remaining 18\% arises from format non-compliance, where the model fails to follow the three-step reasoning paradigm. Annotation guidelines and more details are included in Appendix C.

\section{Conclusion}
We propose MASIR, a reasoning paradigm that abstracts spatial problems into mathematical form and solves them through three steps: Problem Abstraction, Spatial Perception, and Mathematical Computation. To train this capability, an automated data construction pipeline and a staged training scheme are developed. Experiments across six benchmarks demonstrate consistent and substantial improvements over base models. Ablation studies further confirm that the gains stem primarily from the Problem Abstraction stage, isolating the contribution of mathematical representation from the training structure. Error analysis identifies Spatial Perception as the dominant failure source, indicating that perception capability is the primary bottleneck, and that stronger perception would directly translate to further gains.

\bibliography{aaai2027}


\end{document}